\documentclass{article}
\usepackage[nonatbib,final]{neurips_2020}
\usepackage[utf8]{inputenc}
\usepackage{hyperref,algorithm}
\usepackage{algorithmic}
\usepackage{graphicx}
\usepackage{amsmath,amssymb,amsthm}
\usepackage[dvipsnames]{xcolor}
\usepackage{flushend,wrapfig,epsfig}
\usepackage{subcaption}
\newcommand{\tc}[1]{{\color{black}#1}}

\usepackage{bm}

  \providecommand{\cD}{\mathcal{D}}

  \providecommand{\cN}{\mathcal{N}}

\newcommand{\abs}[1]{\left\lvert #1\right\rvert}
\newcommand{\norm}[1]{\left\lVert #1\right\rVert}

\DeclareMathOperator{\argmin}{arg\,min}

\newcommand{\set}[1]{\{#1\}}

\def\remark{\addtocounter{remark}{1}\def\@currentlabel{\theremark}%
\emph{Remark~\theremark}. } \makeatother
\newcounter{remark}

\def\figref#1{Figure~\ref{#1}}

\def\algref#1{Algorithm~\ref{#1}}

\newtheorem{claim}{Claim}

\newcounter{mycounter}  
\newenvironment{noindlist}
 {\begin{list}{\arabic{mycounter}.~~}{\usecounter{mycounter} \labelsep=0em \labelwidth=0em \leftmargin=0em \itemindent=0em}}
 {\end{list}}
\title{Hybrid Federated Learning:\\ Algorithms and Implementation}
\author{
Xinwei Zhang\\
University of Minnesota\\
\texttt{zhan6234@umn.edu}
\And
Wotao Yin\\
University of California, Los Angeles\\
\texttt{wotaoyin@math.ucla.edu}\\
\And 
Mingyi Hong\\
University of Minnesota\\
\texttt{mhong@umn.edu}
\And 
Tianyi Chen\\
Rensselaer Polytechnic Institute\\
\texttt{chent18@rpi.edu}
}
\date{April 2020}

\begin{document}

\maketitle
\vspace{-0.3cm}
\begin{abstract}
\vspace{-0.3cm}
    Federated learning (FL) is a recently proposed distributed machine learning paradigm dealing with distributed and private data sets. Based on the data partition pattern, \tc{FL} is often categorized into horizontal, vertical, and hybrid settings. Many works have been developed for the first two settings. The hybrid FL setting, which deals with partially overlapped feature space and sample space, remains \tc{less} explored, though this setting is extremely important in practice. 
    In this paper, we first set up a new model-matching-based problem formulation for hybrid FL and then propose an efficient algorithm that can collaboratively train the global and local models to deal with full and partial featured data. We conducted numerical experiments on the multi-view ModelNet40 data set to validate the performance of the proposed algorithm. {To the best of our knowledge, this is the first  formulation and algorithm developed for the hybrid FL.} 
    \vspace{-0.3cm}
\end{abstract}

\section{Introduction}
\vspace{-0.3cm}
Federated learning (FL) is an emerging machine learning framework where multiple clients --- such as mobile devices or organizations --- collaboratively train a machine learning (ML) model \cite{mcmahan2017}. FL specifically addresses the new challenges including the difficulty of synchronizing multiple clients, the heterogeneity of data, and the privacy and security of clients' data. When their local models are not Private, they may also need to be protected. Due to these challenges, classic ML methods cannot be directly applied  \cite{kairouz2019}. 

A popular FL setting that partitions data among clients is called \textit{horizontal FL} (HFL). Each client has the data of a different set of subjects, and the data of every client have the same set of features \cite{konevcny2016federated, li2020FLChallenges}. Examples of such data include smartphone users' word-typing histories \tc{(from the same word dictionary)}, which are stored on individual devices and analyzed by the same features \cite{mcmahan2017}. 
One can apply HFL to learn a model for word or sentence completion.
In the setting known as \textit{vertical FL} (VFL), it is features that are partitioned among clients, and all the clients share a common set of subjects~\cite{chen2020vafl,liu2019communication}. More features help build a more accurate model than using fewer features.
For example, VFL can help an insurance company better predict someone's risk using not just this person's records at this company but also his/her records from multiple other insurance businesses. 

Besides having different training approaches, HFL and VFL also have different prediction processes. 
In HFL, the clients share the jointly-trained model, so each client performs predictions independently. 
In VFL, while a client can predict using its local model based on its local features, more accurate predictions are made when more clients work together and use their jointly learned model that takes all the features available.




\noindent{\bf Motivation and main challenges.} In practice, however, a client may possess only some subjects and some features. It is possible that no client has all the features or all the subjects. 
This is the case of financial institutions such as insurance providers, banks, and stock services, which serve just a fraction of all customers and have only their partial records.
This setting has been referred to as \emph{hybrid FL}~\cite{konevcny2016federated,li2020FLChallenges}, and it is the setting we focus in this paper. 
Both HFL and VFL are special cases of hybrid FL. Compared to HFL and VFL, hybrid FL has its unique challenges.
Some specific ones pertaining to algorithm design are:

\begin{noindlist}
\vspace{-0.2cm}
    \item {\bf Local and global models.} In Hybrid FL training, each client has its local data over a subset of features. 
During inference, however, a piece of input data may have just these features or, possibly, more or even full features.
    In the former case, the client shall make inference by itself.
    In the latter case, however, the client will rely on the server to take the inference.
    So the server is  required to maintain a model that supports all the features.
    
    \item {\bf Limited data sharing.} In typical HFL, the clients do not share their local data or labels. In VFL, the labels are either made available to the server~\cite{chen2020vafl} or stored in a designated client~\cite{liu2019communication}. A Hybrid FL system needs to deal with both types of clients, so it is desirable that the training method can run without requiring the server to access any data, including the labels. 
    
    \item {\bf Sample synchronization.} A typical issue with VFL (where each client has some features of all training samples) is that 
    all the clients need to draw the same mini-bath of samples; this problem is exacerbated in the hybrid FL system because not all the clients have all the samples. An ideal algorithm shall work without requiring the clients to synchronize their sample draws.
    \vspace{-0.2cm}
\end{noindlist}

All the above points will become specific challenges when we design our optimization algorithms, and it will become clear that none of the existing FL methods can meet all these requirements. 

{\bf Related work on HFL.}
In HFL, a common algorithm is FedAvg~\cite{konevcny2016federated}, which adopts the computation-then-aggregation strategy. The clients locally perform a few steps of model updates, and then the server aggregates the updated local models and averages them before sending the updated global model back. Beyond model communication, MIME~\cite{karimireddy2020mime} and SCAFFOLD~\cite{karimireddy2019scaffold} also send local gradients and other statics to the server to achieve better convergence. Furthermore, PNFM~\cite{yurochkin2019PFNM} and FedMA~\cite{wang2020fedMA} use a parameter-matching-based strategy in place of the model averaging step to get better global model performance, and they do not require the global model to have the same size as the local models. All HFL algorithms assume their data have the same size and format.

{\bf Related work on VFL.}
In VFL, the features and thus the models are divided over different clients~\cite{hardy2017private, ma2019privacy, liu2019communication, chen2020vafl}. VFL has fewer results than HFL. Federated Block Coordinate Descent (FedBCD)~\cite{liu2019communication} uses a parallel BCD-like algorithm to optimize the local blocks and transmits essential information for the other clients to compute their local gradients. Vertical Asynchronous Federated Learning (VAFL)~\cite{chen2020vafl} assumes that the server holds the global inference model while local clients train the feature extractors that deal with the local features. 

\noindent{\bf Our contributions.} We summarize our main contributions as follows. 
\begin{noindlist}
	\vspace{-0.15cm}
\item We propose a hybrid FL model that addresses many key needs of collaborative-learning scenarios, where neither the subject set nor the feature set is necessarily complete at any client. Such a formulation can be  tailored to meet the requirements in different FL settings. To our knowledge, this is the first concrete hybrid FL model in the literature.

	\vspace{-0.15cm}
\item We develop a convergent hybrid FL algorithm that enables knowledge transfer among clients. At the same time, our approach {maintains data locality} 
and improves communication efficiency by removing the requirement of sample synchronization . 

	\vspace{-0.15cm}
\item We evaluated the performance of the hybrid FL algorithm on a real dataset. The learned model achieved an accuracy that was comparable to that by a centrally trained model.  
\end{noindlist}
\vspace{-0.3cm}





\section{Problem Formulation}
\begin{figure}[htb]
\def\epsfsize#1#2{0.25#1}
\centerline{\epsffile{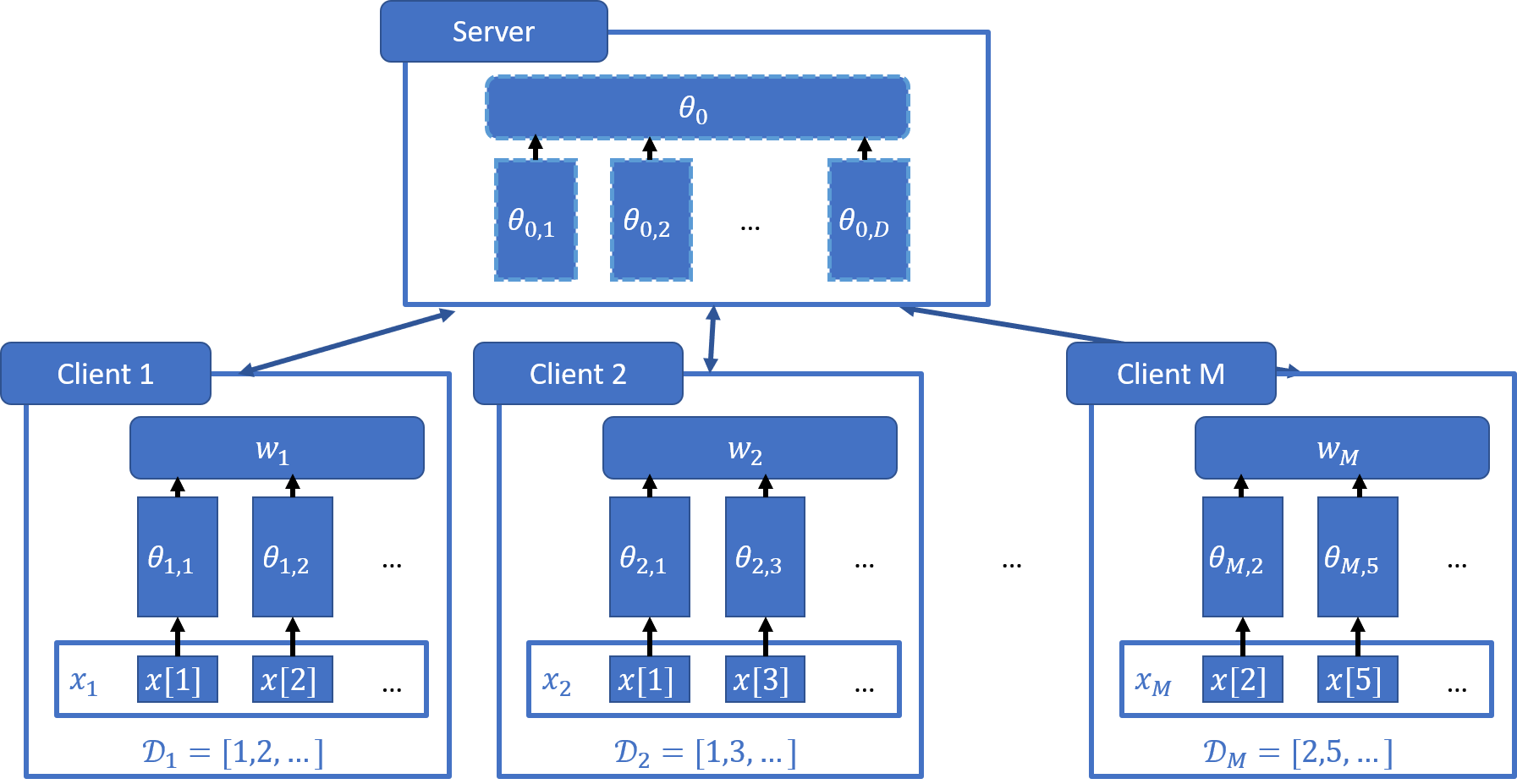}}
  \caption{The proposed Hybrid FL system structure}
\label{fig:system}
\vspace*{-0.5cm}
\end{figure}
Consider $N$ samples, written as $\{\mathbf{x}^n, y^n\}_{n=1}^N$, where each $\mathbf{x}^n$ has $D$ features $\mathbf{x}^n = [\mathbf{x}[1]^n;\cdots; \mathbf{x}[D]^n]$ and $y^n$ is the label. 
There are $M$ clients, each \tc{having} some samples and their partial features.
Index the features by $d=1,\dots,D$, sample data by $n=1,\dots,N$, and clients by $m=1,\dots,M$. If the $m$th agent has the $n$th sample, we write $\mathbf{x}_m^n = [\mathbf{x}[d_1]^n;\cdots; \mathbf{x}[d_m]^n]_{d_1,\dots,d_m\in \cD_m}$, {where $\mathcal{D}_m$ is the set of features available to the $m$th agent.} 

\noindent {\bf A Generic Formulation.} Consider a hybrid FL model consisting of an inference model and some feature extractors. 
For a given agent $m$, if it has enough samples 
that are related to feature $d$, then it will learn a {\it feature extractor} $\theta_{m,d}$. If an agent $m$ does not have enough sample containing a feature $d'$, then it will not participate in learning $\theta_{m,d'}$. There is also a global feature extractor $\theta_{0,d}$ at the server, and  $\theta_{0,d}\approx \theta_{m,d} $ for every agent $m$ that has its local $\theta_{m,d}$.

Each agent $m$ processes an input sample $\mathbf{x}^n_m$ by going through an embedding $h_m(\{\theta_{m,d}\}_{d\in {\cal D}_m}; \mathbf{x}_m^n)$.
Its result, \tc{an embedding vector}, goes through the inference model parameterized by $w_m$, also learned by agent $m$, and produces a label output. 
The server creates global models $\theta_0$ and $\{\theta_{0,d}\}_{d=1,\dots,D}$, where $\{\theta_{0,d}\}_{d=1,\dots,D}$ approximates the local models $\{\theta_{m,d}\}_{d\in \mathcal{D}_m}$, $m\in[M]$ and $\theta_0$ represents the model that maps the outputs of $\{\theta_{0,d}\}_{d=1,\dots,D}$ to a label. The setting discussed above is illustrated in Figure~\ref{fig:system}.




Consider a high-level problem formulation as below: 
    \begin{align}\label{eq.hyFL}
        \min_{\{\{\theta_{m,d}\}, w_m\}}~~ &\frac{1}{MN}\sum^M_{m=1}\sum_{n\in \cN_m} \ell \left(h_m(\{\theta_{m,d}\}; \mathbf{x}_m^n);w_m; y^n\right)+ r(\{\{\theta_{m,d}\}, w_m\}; \theta_0, \{\theta_{0,d}\})
    \end{align}  
where $\ell(\cdot)$ measures the accuracy of using the embedding $h(\{\theta_{m,d}\}_{d\in {\cal D}_m}; \mathbf{x}_m^n)$ and the local model $w_m$ to predict $y^n$, and $r(\cdot)$ is a generic regularizer that encodes the prior knowledge about the global and local models.  Such a formulation is general enough to capture several special Hybrid FL settings. {It is easy to see that the exiting VFL and HFL settings are both its special cases. Next, we demonstrate how to customize problem \eqref{eq.hyFL} to a specific hybrid FL problem.}

\noindent {\bf A Feature Matching Based Hybrid FL Formulation.} In this specific setting, we assume that both features and labels are {\it private}, and they are not shared with other clients nor with the server. Also assume that the \tc{feature extractors for the same feature are approximately consensual}, that is, $\theta_{m,d} \approx \theta_{0,d}$ for every agent $m\in[M]$. In this case, we let $\theta_m$ denote the concatenates $\{\theta_{m,d},\, d\in {\cal D}_m\}$ and $\{\mathbf{x}_m^n, y^n\}_{n\in {\cal N}_m}$ denote the local data set. Then the objective function of \eqref{eq.hyFL} can be separated into the sum of $M$ local objectives, each of which is:
\begin{equation}\label{eq.hyFL-1}
\begin{aligned}
\min_{\{\theta_{m,d}, w_m\}}~~~&\frac{1}{N}\sum_{n\in{\cal N}_m}\ell \left(h_m(\theta_m; \mathbf{x}_m^n);w_m; y^n\right)+ r_m(\theta_m, w_m;\theta_0, \{\theta_{0,d}\}).
\end{aligned}
\end{equation}

Here $r_m(\cdot)$ indicates the local regularizer for client $m$ that \tc{enforces} the consensus among the local feature extractors $\theta_m$ and regularizes the difference of the local inference model $w_m$. Our main design effort will be devoted to finding the proper regularizer $r_m(\cdot)$, 
{which has the following desired features: 1) It helps enforce the consensus of $\theta_{m,d}$, $m\in M$ with $d\in \mathcal{D}_m$, and $\theta_{0,d}$; 2) It facilitates the learning of a global inference model $\theta_0$ from the local inference models $\{w_m\}_{m=1}^{M}$.}
Since these two tasks are relatively separable, it is natural to express $r_m(\cdot)$ as : 
\begin{align}\label{c}
r_m(\theta_m, w_m;\theta_0, \{\theta_{0,d}\}) = \sum_{d\in \mathcal{D}_m} r_{1}(\{\theta_{m,d}\}, \{\theta_{0,d}\}) +r_{2}(\{w_m\},\theta_0).
\end{align}
We can use any reasonable distance function to construct $r_1(\cdot)$ since $\theta_{m,d}$ and $\theta_{0,d}$ have the same dimension. However, it is not straightforward to construct $r_2(\cdot)$ for the following reasons. First,  the input dimension of $\theta_{0}$ is much larger than each individual $w_m$ since $\theta_{0}$ is the inference model that takes all the features extractors as inputs. Second, it is not easy to identify the relationship between different $w_m$'s parameters and combine them to yield a global $\theta_0$. 

To deal with these challenges, we adopt a {\it matched averaging} idea proposed in~\cite{yurochkin2019PFNM, wang2020fedMA}, which is used in the HFL setting to dynamically match the neurons of local models to build a global model. More specifically, it is assumed that the global $\theta_0$ and local $w_m$ are related through a linear mapping $w_m \approx \Pi_m \theta_0$, where such a mapping $\Pi_m$ should be optimized.  One special case of $\Pi_m$ is a matching matrix containing only one non-zero entry at each row. Using such a matching matrix ensures that each neuron in the global model is a linear combination of a set of most closely related neurons in the local models. %
While the idea of model matching has been explored in the recent works  \cite{yurochkin2019PFNM, wang2020fedMA}, here we design a special matching strategy specifically for hybrid FL. 


By using the model matching strategy, the two regularizers in \eqref{eq.hyFL} can be expressed as following:
\begin{align}
r_{1}(\{\theta_{m,d}\}, \{\theta_{0,d}\}) = 
{\rm dist}(\theta_{m,d},\theta_{0,d}), \quad r_{2}(\{w_m\},\theta_0) = 
{\rm dist}(w_m,\Pi_m\theta_{0}),
\end{align}
where $\rm dist(\cdot)$ measures the distance between the models. It is important to note that the matching patterns $\Pi_m$'s have to be optimized as well.  
This leads to the following  Hybrid FL problem: 
\begin{equation}\label{eq.hyFL-1main}
\begin{aligned}
\min_{\{\theta_{m}, w_m, \Pi_m\}, \theta_0, \{\theta_{0,d}\}}~~~&\frac{1}{MN} \sum_{m=1}^M \sum_{n\in{\cal N}_m}\ell \left(h(\theta_m; \mathbf{x}_m^n);w_m; y_m^n\right)\\
&+ \mu\sum_{m=1}^M {\rm dist}(\Pi_m\theta_0, w_m) + \sum^M_{m=1}\sum_{d\in\cD_m}{\rm dist}(\theta_{m,d},\theta_{0,d})\\
\mbox{s.t.}~~~& \Pi_m\mathbf{1} = \mathbf{1}, \; \Pi_m\ge 0,\; m=1,\dots, M.
\end{aligned}
\end{equation}
In this formulation, we jointly minimize the classification loss and the consensus loss and use $\mu$ to balance between the two losses. When $\mu$ is very small, the clients focus on training the local models, which can be quite different across the clients. When $\mu$ is large, the emphasis is put on learning an accurate global model by integrating local information. 



\section{Proposed Algorithm}

We propose an algorithm for solving \eqref{eq.hyFL-1main}. The problem contains parameter blocks 
$\{\theta_{m,d}\}$ and $\{w_m\}$, the global parameters $\theta_{0,d}$, and the global parameters $\theta_{0}$ and $\{\Pi_m\}$. We can update each of them given the others. 
The problem related to the local parameters $\theta_m, w_m$, $m=1,\dots,M$, is:
\begin{equation}\label{eq:local_prox}
    \min_{\theta_{m}, w_m} f_m(\theta_{m}, w_m) = \sum_{n\in{\cal N}_m}\ell \left(h(\theta_m; \mathbf{x}_m^n);w_m; y_m^n\right) + \mu {\rm dist}(\Pi_m\theta_0, w_m) + \sum_{d\in\cD_m}{\rm dist}(\theta_{m,d},\theta_{0,d});
\end{equation}
the problem related to global feature extractors $\theta_{0,d}$'s is:
\begin{equation}\label{eq:feature_ex}
    \min_{\theta_{0,d}}\sum_{m=1}^M\sum_{d\in\cD_m}{\rm dist}(\theta_{m,d},\theta_{0,d});
\end{equation}
and the third block related to the global inference model $\theta_0, \Pi_m$ is:
\begin{equation}\label{eq:matching}
    \min_{\theta_0, \set{\Pi_m}}\sum_{m=1}^M {\rm dist}(\Pi_m\theta_0, w_m),~~\mbox{s.t.}\;\Pi_m\mathbf{1} = \mathbf{1},\;\Pi_m\ge 0,\; m=1,\dots, M.
\end{equation}
\tc{In view of its block structure}, we propose a block coordinate descent type algorithm called Hybrid Federated Matched Averaging (HyFEM) to solve this problem, which is summarized in \algref{alg:HyFEM}. 

\begin{algorithm}[t]
\begin{algorithmic}
\small
	\STATE {{\bfseries Input:} $\set{w^0_m}, \set{\theta^0_d}, \eta, T, Q, P$}\\
	\STATE {Initialize: $\theta^0_{m,d}= \theta^0_d,~m=1,\dots,M$}
	\FOR{$t=0,\dots,T-1$}
	    \vspace{2pt}
	    \STATE \mbox{// \emph{Local Updates}}:\\
		\FOR{$i=1,\dots,M$ in parallel}
			\STATE $\theta^{r,Q}_{m}, w_m^{r,Q} = \argmin_{\theta_{m}, w_m} f_m(\theta_{m}, w_m)$
        \ENDFOR
        \vspace{2pt}
        \STATE \mbox{// \emph{Global Aggregation of Feature Extractors}}:\\
        \STATE $\theta^{t+1}_d = \frac{1}{\abs{\set{m:d\in\cD_m}}}\sum_{m:d\in\cD_m}\theta^{t,Q}_{d,m}$, \quad {for $d=1,\dots, D$}\\
        \vspace{2pt}
        \STATE \mbox{// \emph{Global Matching Updates}}:\\
        \FOR{$p = 0,\dots,P-1$}
            \STATE Randomly select a client index $m'$
            \STATE Update matching pattern: $\Pi_{m'}= \argmin_{\Pi_{m'}}f_0(\theta^{t,p}_0,\set{\Pi_m};\set{w^{t,Q}_m})$ 
            \STATE Update global inference model: $\theta^{t,p+1}_0 = \argmin_{\theta_0}f_0(\theta_0,\set{\Pi_m};\set{w^{t,Q}_m})$
        \ENDFOR
        \STATE Distribute the aggregated models: $w^{t+1,0}_m = \Pi_m\theta^{t,P}_0,~\theta^{t+1,0}_{m,d} = \theta^{t+1}_d$
    \ENDFOR
\end{algorithmic}
\caption{Hybrid Federated Matching Algorithm}\label{alg:HyFEM}
\end{algorithm}

In each iteration, the local clients first fix $\Pi_m$ and $\theta_0$ while optimizing the local objective function \eqref{eq:local_prox} with gradient-based local solvers such as SGD. The form of this local problem is similar to FedProx \cite{li2020FLChallenges}, so we call it HyFEM-Prox. Then the server aggregates the local feature extractors and optimizes \eqref{eq:feature_ex}. For a common choice of the distance function, such as the square of the Euclidean norm, the problem has a closed-form solution. 
Finally, the server aggregates the local models and optimizes the global model matching problem~\eqref{eq:matching}. This subproblem can be optimized by another iterative procedure: a) randomly pick an index $m$ and apply the Hungarian matching algorithm to find $\Pi_m$ fixing the other $Pi_{m'}$'s; b) update $\theta_0$ by fixed $\set{\Pi_m}$. After a few rounds of updates, we obtain the matched global model and the corresponding matching pattern. When the parameters represent deep neural networks, the matching is performed layer by layer. Dummy neurons with zero weights are padded to match the size of different models. Due to space limitation, we skip some technical details.

{We have the following convergence results about the proposed algorithm.
\begin{claim}
Suppose that for each $m\in\{1,2,\cdots, M\}$, $f_m$ has Lipschitz continuous gradients w.r.t. $\theta_m$ and $w_m$; further assume that $\mbox{dist}(\cdot, \cdot)$ has Lipschitz gradients w.r.t. each of its argument. Suppose that $\theta_0$ has a fixed dimension and its size is bounded. Then Algorithm 1 converges to a first-order stationary solution for problem \eqref{eq.hyFL-1main}.
\end{claim}

This claim can be proved by observing that the proposed algorithm can be viewed as the classical block-coordinate gradient descent (BCGD) algorithm, so the classical results~\cite{zeng2019global,razaviyayn2013unified} can apply. It can also be extended to the case where the $\theta_m$ and $w_m$ steps are not solved to global minima but to some approximate stationary solutions of \eqref{eq:local_prox}. Due to space limitation, we omit the detailed proofs.}

\begin{remark}  We highlight the merits of the proposed approach:
1) Unlike the typical VFL formulations \cite{liu2019communication,chen2020vafl}, our approach keeps the labels at the clients. Hence, the local problems are fully separable. There is no sample-drawing synchronization needed during local updates; 
2) By utilizing the proposed merging technique, we can generate a global model at the server, which makes use of full features. {This makes the inference stage flexible: the clients can use either partial features (by using its local parameters $(\{\theta_{m,d}\}, w_m)$) or the full features by requesting $(\{\theta_{0,d}\}, w_0)$ from the server or letting the server do the inference}.
\end{remark}

\begin{remark} Our formulation \eqref{eq.hyFL-1main} and our proposed algorithm naturally reduce to existing FL models. Consider, for example, every local client has all the features and their matching patterns are fixed, and the distance function is chosen as $\ell_2$-norm square (i.e., ${\rm dist}(a,b)=\norm{a-b}^2_2$). Then, HyFEM becomes FedProx with the local problem~\eqref{eq:local_prox}.
\end{remark}

\begin{remark} 
In practice, we inexactly solve the local problem~\eqref{eq:local_prox}.
As an alternative, locally we can ignore the regularizer and approximately solve the following local problem for a few iterations:
\begin{equation}\label{eq:local_avg}
    \min_{\theta_{m}, w_m} f_m(\theta_{m}, w_m) = \sum_{n\in{\cal N}_m}\ell \left(h(\theta_m; \mathbf{x}_m^n);w_m; y_m^n\right).
\end{equation}
We name this alternative as HyFEM-Avg. Compare with HyFEM-Prox, the gradient estimation is easier, and it requires much less memory of the local clients.
\end{remark}

\begin{figure}[h]
    \vspace{-0.3cm}
\centering
\includegraphics[width=0.8\textwidth]{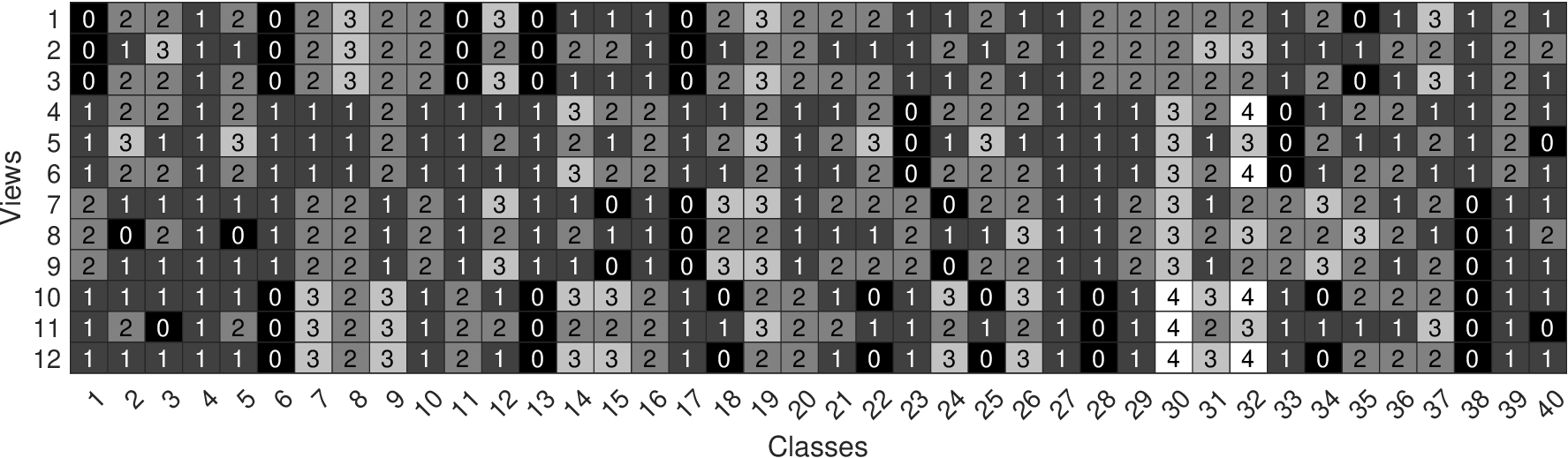}
\vspace*{-0.1cm}
  \caption{\small Illustration of how many clients (the numbers in boxes) possess the training data of each view in each class, in an experiment with $M=8$ clients, $D=12$ views, and 40 classes.}
\label{fig:count}
\vspace{-0.5cm}
\end{figure}


\subsection{Experiment Settings}

To evaluate the proposed algorithms, we conducted experiments using the ModelNet40 data set for multiview object classification. The dateset has 40,000 samples from 40 classes. Each sample consists of 12 views from different angles, which are the 12 features of an object. We used $M = 4, 8$ clients during the experiment, and each client had data from only partial views in some of the classes. We used the convolutional layers of Resnet-34 for feature extraction for HyFEM and used an MLP with one hidden layer for local inference. For comparison, we also trained the model using the entire data set with all the features, which we label ``centralized'' in the figures.
\begin{figure}[t]
\vspace{-1.0cm}
    \centering
    \begin{subfigure}[t]{0.46\linewidth}
        \centering
        \includegraphics[width=\linewidth]{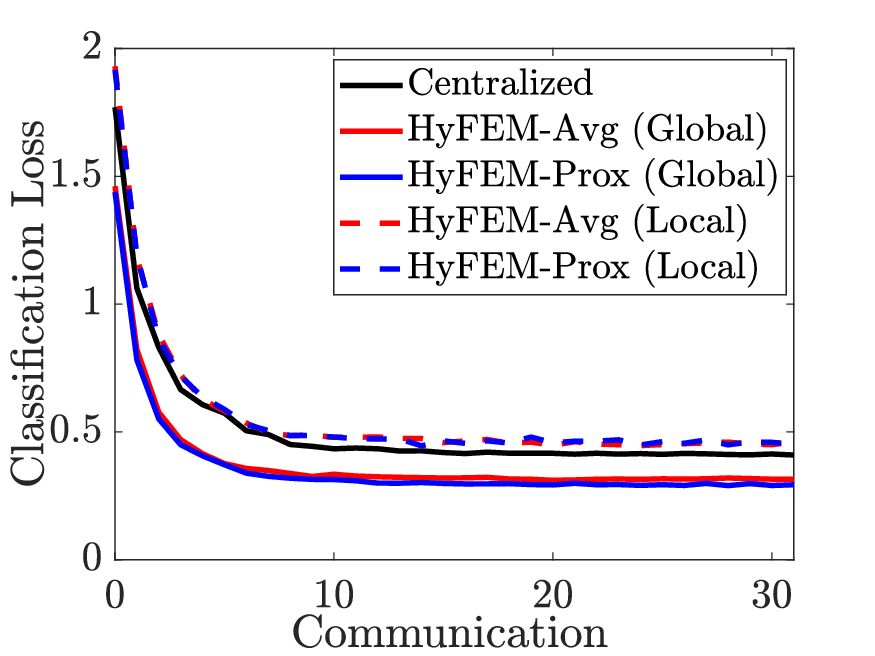}
        \caption{\small Testing losses of Central training, HyFEM-Avg, and HyFEM-Prox ($\mu=0.1$) versus the number of communication rounds.}
    \end{subfigure}
    \hfill
    \begin{subfigure}[t]{0.46\linewidth}
        \centering
        \includegraphics[width=\linewidth]{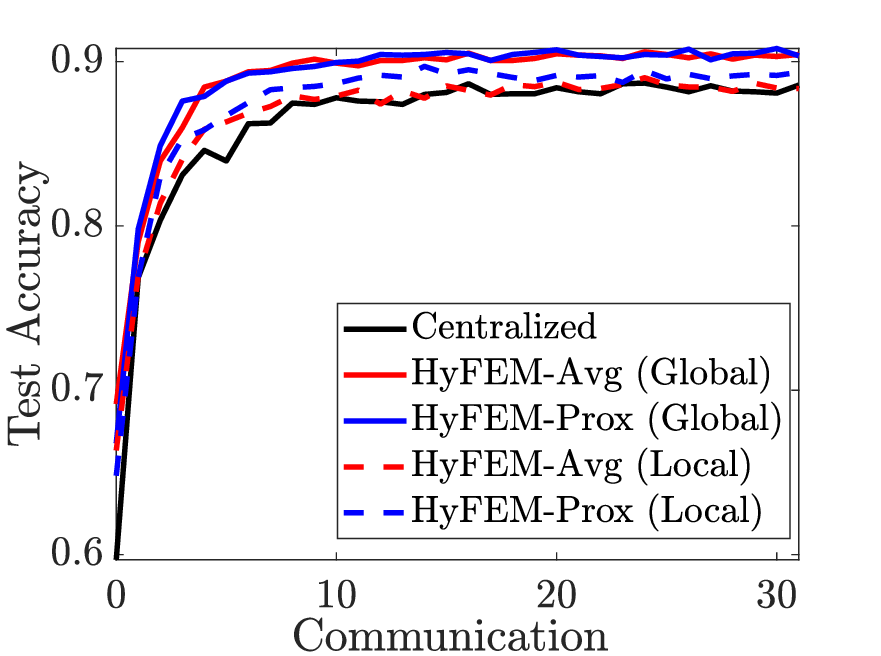}
        \caption{\small Testing accuracies of Central training, HyFEM-Avg, and HyFEM-Prox ($\mu=0.1$) versus the number of communication rounds.}
    \end{subfigure}
    \caption{\small Prediction performance on 4,000 4-view data points from ModelNet40. The neural networks were trained with data points of totally 4 views. In HyFEM, each client had a unique portion of the training data with only 3 of the 4 views.} 
    \label{fig:view_4_3}
\end{figure}
\begin{figure}[t]
    \vspace{-0.3cm}
    \centering
    \begin{subfigure}[t]{0.44\linewidth}
        \centering
    \includegraphics[width=\linewidth]{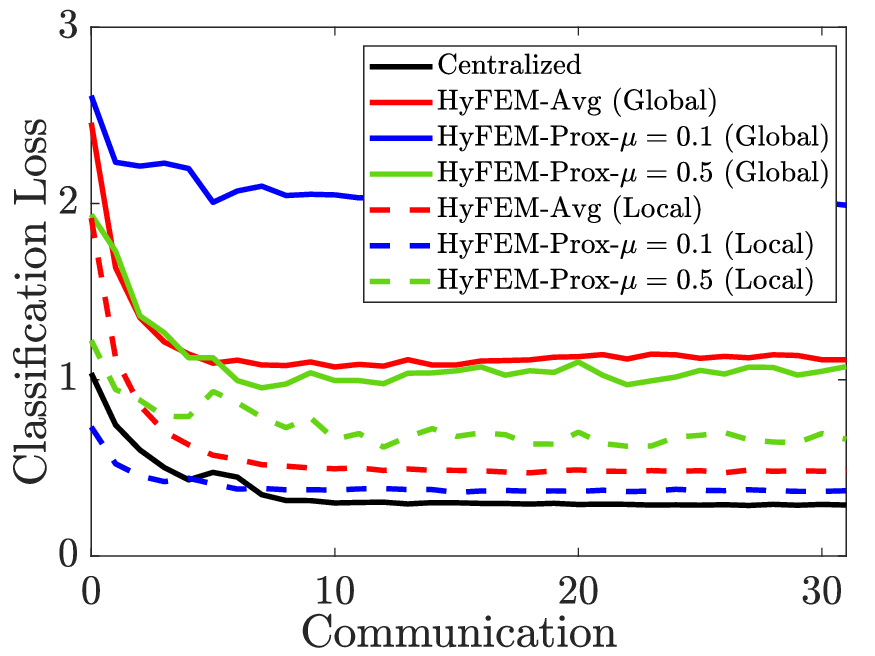}
        \caption{\small Testing losses of Central training, HyFEM-Avg, HyFEM-Prox ($\mu=0.1$), and HyFEM-Prox ($\mu=0.5$) versus the number of communication rounds.}
    \end{subfigure}
     \hfill
    \begin{subfigure}[t]{0.44\linewidth}
        \centering
        \includegraphics[width=\linewidth]{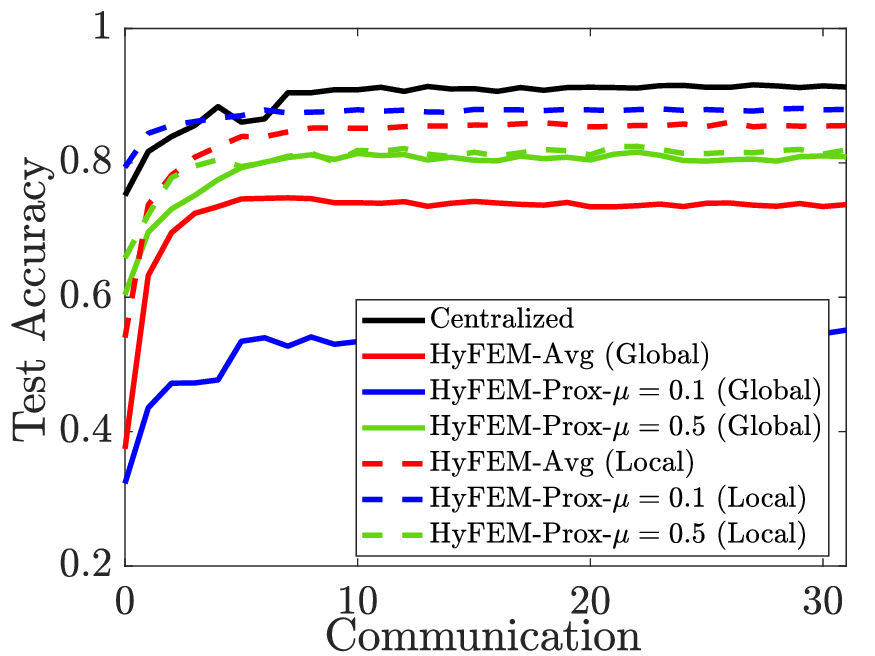}
        \caption{\small Testing accuracies of of Central training, HyFEM-Avg, HyFEM-Prox ($\mu=0.1$), and HyFEM-Prox ($\mu=0.5$) versus the number of communication rounds.}
    \end{subfigure}
    \caption{\small Prediction performance on 4,000 12-view data points from ModelNet40. The neural networks were trained with data points of totally 12 views. In HyFEM, each client had a unique portion of the training data with only 6 out of the 12 views. }
    \label{fig:view_12}
    \vspace{-0.2cm}
\end{figure}

In the training phase, the local clients trained their models with their partial data and partial features along with the server aggregating their local models. In the testing phase, the local clients tested their local models on all the samples but only using their corresponding partial features. We average over all the clients to obtain the averaged local accuracy. The global accuracy is computed using the matched global model $(\theta_0,\{\theta_{m,0}\})$ on all the samples with full features.

We uniformly set the total communication rounds $T = 32$ and local update step $Q = 32$ with HyFEM-Avg with mini-batch size $32$ on the local clients during training. The initial learning rate was set to $\eta = 0.005$ and decayed by $0.2$ for every 8 rounds of communication.

In the test with $M=4$ clients, we selected four views ($D = 4$) as the full set of features. Each client had 30 classes of data and had 
$\abs{\cD_m} = 3$ out of the 4 views. Therefore, none of the clients had either the full samples or the full features. The data distribution was heterogeneous. 
\figref{fig:view_4_3} depicts the result of this test.
The local models and the global model trained with HyFEM had good test results and even obtained a higher accuracy than the model that was centrally trained. We believe that data heterogeneity helped reduce model over-fitting.

The second test was more challenging. It had $M=8$ clients and $D = 12$ views in total. Each client had only $15$ classes and $6$ views of data, and the way we divided data and features is depicted in \figref{fig:count}: $12.08\%$ of the data was never used by any local client during training, and the data distributions were more heterogeneous than the last test with $M=4$. In addition, we set the penalty weight $\mu = 0.1$ and $\mu = 0.5$ for HyFEM-Prox to {understand how this parameter affects the global and the local accuracy}. \figref{fig:view_12} shows the testing losses and accuracies of the algorithms. We can see that the federated models behaved worse than the centrally trained model, which is predictable because of the high data heterogeneity and the missing data. We can also observe that $\mu$ had a balancing effect between the global and the local accuracy. When $\mu$ was small ($\mu=0.1$), the local accuracies were high, and the global accuracy was low; when $\mu$ grew larger, the local accuracies dropped and the global accuracy improved. {This is intuitive since by using a larger parameter $\mu$, we put more emphasis on the global model integration.}

\bibliographystyle{IEEEbib}
\bibliography{references}
\end{document}